\documentclass{article}

\usepackage{PRIMEarxiv}

\usepackage[utf8]{inputenc} % allow utf-8 input
\usepackage[T1]{fontenc}    % use 8-bit T1 fonts
\usepackage{hyperref}       % hyperlinks
\usepackage{url}            % simple URL typesetting
\usepackage{booktabs}       % professional-quality tables
\usepackage{amsfonts}       % blackboard math symbols
\usepackage{nicefrac}       % compact symbols for 1/2, etc.
\usepackage{microtype}      % microtypography
\usepackage{lipsum}
\usepackage{fancyhdr}       % header
\usepackage{graphicx}       % graphics
\graphicspath{{media/}}     % organize your images and other figures under media/ folder
\usepackage{algorithm}
\usepackage{xcolor}         % colors
\usepackage{algorithmic}
\usepackage{amsthm}
\usepackage{amssymb}
\usepackage{threeparttable}
\usepackage{makecell}
\usepackage{subfigure}
\usepackage{amsmath}
\usepackage{multirow}
\usepackage{bm}
\usepackage{bbding}
\usepackage{wrapfig}

\def\benchmarkname{OCDB}
\newtheorem{Example}{Example}
\newtheorem{Lemma}{Lemma}
\newtheorem{Definition}{Definition}
\newtheorem{Proposition}{Proposition}
\title{OCDB: Revisiting Causal Discovery with a Comprehensive Benchmark and Evaluation Framework
}

\author{
  Wei Zhou, Hong Huang\thanks{Hong Huang is the corresponding author. Wei Zhou, Hong Huang, Ruize Shi, Kehan Yin and Yuanyuan Lin are affiliated with the National Engineering Research Center 
for Big Data Technology and System, Services Computing Technology and System 
Lab, Cluster and Grid Computing Lab, School of Computer Science and Technology, 
Huazhong University of Science and Technology, Wuhan, 430074, China.}, Guowen Zhang, Ruize Shi, Kehan Yin, Yuanyuan Lin\\
Huazhong University of Science and Technology, China\\
  \texttt{\{weizhou2021, honghuang, lostgreen, rzshi, kehanyin, linyy\}@hust.edu.cn} \\
  \AND
  Bang Liu \\
  DIRO, Université de Montréal \& Mila \& Canada CIFAR AI Chair, Canada \\
  \texttt{bang.liu@umontreal.ca} \\
}

\begin{document}
\maketitle

\begin{abstract}
Large language models (LLMs) have excelled in various natural language processing tasks, but challenges in interpretability and trustworthiness persist, limiting their use in high-stakes fields. Causal discovery offers a promising approach to improve transparency and reliability. 
However, current evaluations are often one-sided and lack assessments focused on interpretability performance. Additionally, these evaluations rely on synthetic data and lack comprehensive assessments of real-world datasets. These lead to promising methods potentially being overlooked.
To address these issues, we propose a flexible evaluation framework with metrics for evaluating differences in causal structures and causal effects, which are crucial attributes that help improve the interpretability of LLMs. We introduce the {\bf O}pen {\bf C}ausal {\bf D}iscovery {\bf B}enchmark (\benchmarkname{}), based on real data, to promote fair comparisons and drive optimization of algorithms. Additionally, our new metrics account for undirected edges, enabling fair comparisons between Directed Acyclic Graphs (DAGs) and Completed Partially Directed Acyclic Graphs (CPDAGs). Experimental results show significant shortcomings in existing algorithms' generalization capabilities on real data, highlighting the potential for performance improvement and the importance of our framework in advancing causal discovery techniques.
\end{abstract}

% keywords can be removed
% \keywords{First keyword \and Second keyword \and More}

\section{Introduction}
\label{sec:introduction}
In recent years, Large Language Models~(LLMs) have demonstrated outstanding performance in various natural language processing tasks, garnering widespread attention~\cite{GPT-4, gemini}. However, issues of interpretability and trustworthiness remain pressing challenges~\cite{interpretability1, interpretability2}. LLMs, typically trained on massive datasets with complex algorithms, often function as "black boxes", making their internal decision-making processes difficult to understand. This lack of transparency limits their application in high-risk fields like healthcare and finance, and undermines user trust. As a result, causal graphs have gained attention as tools for enhancing the interpretability and trustworthiness of LLMs~\cite{causal4llm1, causal4llm2}. 

Two key attributes of causal graphs for enhancing the interpretability of LLMs are causal structure and causal effect. 
Causal structures clearly show the causal dependencies between variables, helping LLMs more accurately understand how variables interact and influence each other\cite{causal-structure4llm1, causal-effect4llm2}. Moreover, LLMs learn from massive amounts of text and can easily mistake correlation for causation. Causal structures can correct this misunderstanding, enabling the model to grasp causal logic more precisely and improve the accuracy of its explanations. 
Causal effects, on the other hand, provide more detailed and specific explanations based on the causal structure\cite{causal4llm2, causal-effect4llm2}. By measuring the impact of different variables on the output, they can help LLMs identify the most critical factors, understand the relative importance of variables, and make wiser decisions.

It is tough to get the true causal graph, so the feasible approach right now is to use causal discovery methods to learn causal graphs from data~\cite{Gaussian-SHD+SID, OCDaf-CBC+SHD+SID}. By employing methods that generate causal graphs with smaller differences in causal structures and causal effects, more accurate and reliable causal information is provided to LLMs, thereby enhancing their interpretability and credibility.
However, the current evaluation of causal discovery methods is one-sided. Some works \cite{rohekar2023temporal-SHD-C, cai2023causal-f1-score, cheng2024cuts+-AUROC} primarily focus on classification performance and structural differences. Conversely, metrics like SID \cite{SID} and KD~\cite{StagedTree-KD+CID} used to evaluate causal effects either only focus on the impact of structure on causal effects or only consider intervention distribution differences. This narrow focus leads to the oversight or underestimation of many potentially excellent algorithms within the current evaluation system, thereby hindering progress in work that could significantly enhance the interpretability of LLMs.

Moreover, the current benchmarks for causal discovery lack real data or fail to comprehensively include various types of real data. Some benchmarks, like CSuite \cite{CSuite} and CDML \cite{CDML}, only contain synthetic data. The absence of real datasets limits our understanding of these algorithms' actual performance and may lead to errors in practical applications. While synthetic data plays a crucial role in algorithm development and initial validation, it differs significantly from the real data used to train LLMs. Conversely, benchmarks such as CausalTime \cite{CausalTime} and Causal-learn \cite{Causal-learn} only provide a single type of real data. LLMs, trained on diverse data types, require support from causal graphs generated from varied data sources. The lack of comprehensive real data types makes it challenging to thoroughly evaluate and improve causal discovery algorithms, thereby impeding them to support the interpretability of LLMs effectively.

To address these issues, in this paper, we propose a new, flexible, and comprehensive evaluation framework for causal graphs generated by causal discovery methods. 
Specifically, we first analyze causal structures and causal effects through examples and theoretical research to explore and derive metrics for measuring differences in them. Then, based on our analysis and research results, we clarify the core objectives of current causal discovery evaluation metrics for the first time, and categorize them into three types.
Finally, we introduce a new causal discovery benchmark based on real datasets, named Open Causal Discovery Benchmark (\benchmarkname{}). This benchmark encompasses a wide variety of data types, ensuring a comprehensive representation of various complex scenarios and diverse data. 
Additionally, using our proposed new metrics, we achieve fair comparisons between Directed Acyclic Graphs (DAGs) and Completed Partially Directed Acyclic Graphs (CPDAGs). 
The evaluation aims to identify superior and robust methods to enhance the interpretability of LLMs.
Evaluation results indicate that current causal discovery algorithms exhibit weak generalization capabilities on real data across different scenarios, highlighting significant room for performance improvement.

Our contributions can be summarized as follows:
\begin{itemize}
\item {\bf Metrics for interpertability}. 
Through analysis and discussion, we propose two metrics to evaluate the differences in causal structures and causal effects, respectively. These metrics can help select appropriate causal discovery methods, thereby providing LLMs with accurate and reliable causal information, and enhancing their interpretability and credibility.
\item {\bf Comprehensive real-world benchmark}. We introduce the \benchmarkname{}, a benchmark based on real datasets that covers a wide range of data types, ensuring the representation of diverse and complex scenarios. This standardized, open platform promotes further research and development in causal discovery.
\item {\bf Fair and Robust evaluation}. Our new metrics enable fair comparisons between DAG and CPDAG. Experiments show that current algorithms have weak generalization on real data across various scenarios, indicating significant room for improvement.
\end{itemize}

\section{Related work}
\paragraph{Causal discovery benchmarks.}
To fairly evaluate the performance of causal discovery algorithms, many benchmarks have been proposed, 
and detailed information of them is provided in Appendix~\ref{sec:current_benchmark}.
Based on the data types provided, we can categorize these benchmarks into three categories: {\bf 1. Static data-based}, 
they primarily provide static datasets to evaluate the performance of causal discovery algorithms, such as   bnlearn\footnote{https://github.com/erdogant/bnlearn/}, CDT~\cite{CDT}, Py-causal~\footnote{https://github.com/bd2kccd/py-causal}, CSuite~\cite{CSuite}, CIPCaD-Bench~\cite{CIPCaD}, and Causal-learn~\cite{Causal-learn}. 
{\bf 2. Time data-based}. These benchmarks  provide multi-time series or event sequence data for discovering causal relationships over time. 
Examples of time data-based benchmarks include CauseMe~\cite{munoz2020causeme}, CDML~\cite{CDML}, and CausalTime~\cite{CausalTime}.
{\bf 3. Composite data-based}, like gCastle~\cite{gcastle}, they encompass datasets that combine multiple types of data, such as static, time series, or event sequences. 
Although these benchmarks provide support for fair comparisons of causal discovery algorithms, they still lack real data or fail to comprehensively include various types of real data, resulting in potentially incomplete and inaccurate evaluation results.

\paragraph{Evaluation metrics for causal graphs.}
With a large number of algorithms being proposed, more and more metrics are being used to evaluate their performance. 
Based on the evaluation goals, they fall into three groups:
{\bf 1. Structure error-based}, like SHD~\cite{NOFEARS-SHD}, MRE~\cite{graphite-NLL+MRE}, and HD~\cite{HD}, these metrics only consider whether the true causal edge exists or whether its direction is correct.
{\bf 2. Causal effect error-based}, such as KD~\cite{StagedTree-KD+CID}, CBC~\cite{OCDaf-CBC+SHD+SID}, and SID~\cite{SID}, they aim to compare the causal effect differences between real graphs and predicted graphs.
{\bf 3. Classification error-based}, like F1-score, AUC, and FPR, they treat causal discovery as a binary classification problem and measure the prediction performance of causal edges. 
Current metrics based on structural errors only consider topological differences and overlook variations in interpretability. Methods that focus on causal effects typically consider either the structure or the intervention distribution's impact on causal effects, leading to biased outcomes. Finally, although classification metrics can reflect model performance, they can't adequately represent the core structural differences in causal graphs, so they shouldn't be used alone to measure the quality of causal graphs.
For the analysis between the current metrics and the metrics we proposed, please see Appendix~\ref{sec:discuss}.

\section{Metrics for interpretability}

In this section, we focus on two key points to enhance the interpretability of LLMs: measuring causal structure differences and causal effect differences. We explore how these measurements can assess the impact of causal graphs generated by causal discovery algorithms on the interpretability of LLMs. We first discuss the measurement paradigms for these objectives separately. Then, through case studies and theoretical analysis, we propose the Causal Structure Distance (CSD) for calculating causal structure differences and the Causal Effect Distance (CED) for calculating causal effect differences. The entire analysis process is outlined as follows.

\subsection{Causal structure distance}
The comparison of differences between two causal graphs is based on the premise that they have the same set of variables. Under this condition, the differences in causal structure entirely depend on the edges. In a graph with a fixed number of nodes, the maximum possible number of edges is also determined. Therefore, by using different state encodings to represent the edges, the edge structure of the graph can be converted into a fixed-length encoded string.
Inspired by string comparison, for edge structure encoded strings with equal length, we can use the Hamming distance~\cite{Hamming-distance} to calculate their differences.

\begin{figure}
    \centering
    
\includegraphics[width=0.9\textwidth]{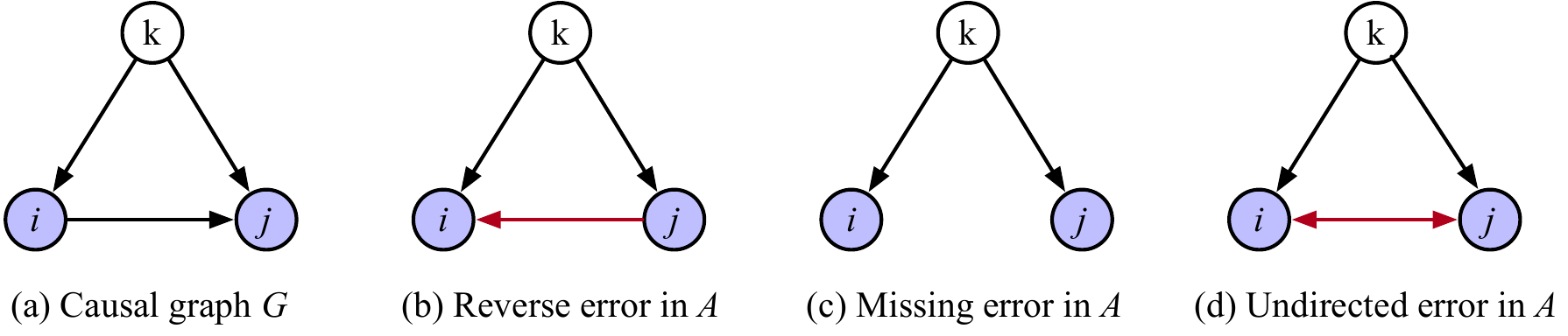}
    
    \caption{A toy example of all incorrect situations on the predicted graph $A$ for the ground truth causal graph $G$.}
    \label{fig:SED-figure}
\end{figure}

\begin{Definition}[Causal Structure Distance]
Let $G$ and $A$ are two causal graphs with the same set of variables $V$, then the Causal Structure Distance(CSD) between them can be defined as
    \begin{equation}
    CSD(G, A) = \sum_{i \in V, j \in V, i\neq j}\mathsf{1}(\mathcal{E}_G^{(i, j)} \neq \mathcal{E}_A^{(i, j)}),
    \label{equation:SED_paradigm}
\end{equation}
where $\mathcal{E}_G$ represents the edge structure encoding of graph $G$, and $\mathcal{E}_A$ denotes the edge structure encoding of graph $A$. $\mathsf{1}(condition)$ is a function that returns 1 when the condition is met and 0 when it is not met.
    \label{Definition:SED_paradigm}
\end{Definition}

However, this definition only considers topological differences and ignores interpretability differences. Different types of structural errors have varying impacts on the interpretability of LLMs, as shown in Example~\ref{Example 1}. Therefore, measuring causal structure differences should account for both topological differences and their impact on interpretability. This approach ensures the selection of more suitable and effective causal discovery algorithms for enhancing the interpretability of LLMs.

\begin{Example}[]
    When using a causal graph to provide interpretability for an LLM, three types of errors can occur with a true causal relationship $i\to j$: reverse, missing, and undirected, as shown in Figure~\ref{fig:SED-figure}(a-d). Reverse errors lead to interpretations opposite to the facts, causing serious decision-making mistakes. Missing errors cause important factors to be overlooked, limiting understanding of the model's behavior. Undirected errors increase interpretation complexity but can still provide useful information. Thus, reverse errors are more serious than missing and undirected errors.

    When there is no causal relationship between $i$ and $j$ in the true causal graph, three types of errors can occur: forward, reverse, and undirected. Forward and reverse errors both seriously mislead interpretation and decision-making. Undirected errors not only mislead decisions but also add complexity and uncertainty to the interpretation. In this case, undirected errors are more severe than forward and reverse errors.
    \label{Example 1}
\end{Example}

Therefore, to better measure the causal structure differences between causal graphs, we need a more reasonable encoding method for edge states. After multiple attempts and considerations, we find that more precise encoding of edges can effectively account for differences in interpretability. For any two distinct nodes $i$ and $j$, instead of using a coarse single-character notation to represent the edge between them, we employ a double-character notation. The first character indicates the state of the directed edge $i \to j$, and the second character indicates the state of the directed edge $j \to i$, where $1$ denotes existence and $0$ denotes non-existence. Furthermore, undirected edges are encoded as existing in both directions simultaneously.

\begin{Example}[]
    As shown in Figure~\ref{fig:SED-figure}, when a directed edge $i \to j$(encoded as 10) actually exists in the graph, the encoding distance of missing(00) and undirected(11) is 1, while the reverse(01) has an encoding distance of 2. A similar conclusion can be drawn for $j \to i$. Conversely, when there is no edge between $i$ and $j$(encoded as 00), the encoding distance of undirected(11) is 2, while the distance of forward(10) and reverse(01) is 1.
    \label{Example 2}
\end{Example}

In this way, we successfully incorporate interpretability significance into causal structure differences. This encoding method is, in fact, the adjacency matrix of a graph. The adjacency matrix encoding not only effectively measures the causal structure differences between causal graphs but also reduces computational overhead. On the one hand, in the adjacency matrix, the distance between two values (0 and 1) allows the use of the absolute value of their difference to replace conditional checks. On the other hand, modern computer systems have optimized matrix operations, significantly enhancing the efficiency of matrix computations.

\begin{Definition}[Matrix Form of CSD]
Let $G$ and $A$ are two causal graphs with the same set of variables, then the causal structure distance between them can be defined as
    \begin{equation}
    CSD(G, A) = \Vert\mathbf{G} - \mathbf{A}\Vert_1,
    \label{equation:SED}
\end{equation}
where $\mathbf{G}$ is the adjacency matrix of $G$, and $\mathbf{A}$ represents the adjacency matrix of  $A$.
    \label{Definition:SED}
\end{Definition}

\subsection{Causal effect distance}
When the causal effects of the same variable differ in two causal graphs, research and analysis based on the incorrect causal graph can lead to misleading conclusions, affecting the interpretability and credibility of LLMs. Therefore, accurately measuring the differences in causal effects is crucial, as it helps in selecting more effective causal discovery algorithms, generating higher-quality causal graphs, and improving interpretability when combined with LLMs. The paradigm for measuring differences in causal effects is similar to that for measuring differences in causal structures.

\begin{Definition}[Causal Effect Distance]
Let $G$ and $A$ are two causal graphs with the same set of variables $V$, then the Causal effect distance(CED) between them can be defined as
    \begin{equation}
    CED(G, A) = \sum_{i\in V, j\in V, i \neq j}\mathsf{1}(CE(i,j)_G \neq CE(i,j)_A).
    \label{equation:SED_1}
\end{equation} 
\end{Definition}

Computing the causal effect is challenging and time-consuming. We require an alternative approach to compare the differences between causal effects in a simpler and more efficient manner.

\begin{Definition}[Reachability Matrix]
Let $G$ is a graph, the reachability matrix can be defined as
    \begin{equation}
    \mathcal{G} = \mathbb{I}(\mathbf{G + I})^r,
    \label{equation:reach}
\end{equation}
where $r$ is equal to $N_v - 1$, $N_v$ is the number of nodes in $G$, and operation $\mathbb{I}$ will set all elements greater than 0 in the matrix to 1.

\end{Definition}

In the reachability matrix, for any two distinct nodes $X$ and $Y$, $\mathcal{G}(X, Y) = 1$ indicates that there is at least one directed path from node $X$ to node $Y$ in graph $G$, implying that $Y$ is a descendant of $X$. According to \cite{pearl}, we have 
\begin{equation}
    \begin{aligned}
p_G(y|do(X=\hat{x})) = \sum_{c\in C}p(y|\hat{x}, c)p(c)
\end{aligned}.
    \label{equation:Intervention1}
\end{equation}
Conversely, $\mathcal{G}(X, Y) = 0$ denotes that there is no directed path between $X$ and $Y$, meaning $Y$ is not a descendant of $X$, then
\begin{equation}
    \begin{aligned}
p_G(y|do(X=\hat{x})) = \sum_{c\in C}p(y|c)p(c)
\end{aligned},
    \label{equation:Intervention2}
\end{equation}
where $C$ is the set of confounding variables, also known as the valid adjustment set.

\begin{Lemma}[Causal Effect Estimation]
Let $G$ is a causal graph, for any two distinct nodes $i$ and $j$, if $\mathcal{G}(i, j) = 1$, then the causal effect from $i$ to $j$ is 
\begin{equation}
\begin{aligned}
    CE(i, j)_G & = p_G(j|do(i=1)) - p_G(j|do(i=0)) \\
    & = \sum_{c\in C}[p_G(j|do(i=1), c) - p_G(j|do(i=0), c)]p(c) \\
    & \neq 0
\end{aligned}.
\label{equation:causal effect1}
\end{equation}
If $\mathcal{G}(i, j) = 0$, then
\begin{equation}
\begin{aligned}
    CE(i, j)_G & = p_G(j|do(i=1)) - p_G(j|do(i=0)) \\
    & = \sum_{c\in C}[p_G(j|c) - p_G(j|c)]p(c) \\
    & = 0
\end{aligned},
\label{equation:causal effect2}
\end{equation}
where $do(i=1)$ represents an intervention on variable $i$, while $do(i=0)$ denotes no intervention. 
\label{Lemma:causal effect}
\end{Lemma}

According to Lemma~\ref{Lemma:causal effect}, we can analyze various scenarios of causal effect comparisons between causal graphs, thereby simplifying the computation of CED. On the one hand, if $\mathcal{G}(i, j) = 1$ and $\mathcal{A}(i, j) = 0$, then it follows that $CE(i, j)_G \neq CE(i, j)_A$. The same conclusion is reached if $\mathcal{G}(i, j) = 0$ and $\mathcal{A}(i, j) = 1$. On the other hand, when $\mathcal{G}(i, j) = \mathcal{A}(i, j)$, whether $CE(i, j)_G$ equals $CE(i, j)_A$ depends on whether the valid adjustment sets are identical. In other words, it is only necessary to verify whether the valid adjustment set $C$ in $A$ remains a valid adjustment set in $G$. Fortunately, Shpitser et al.\cite{shpitser} have provided a rapid and efficient method for testing valid adjustment sets.

\begin{Lemma}[Characterization of valid Adjustment Sets]
Given a DAG or CPDAG $G=\{V, E\}$, for any two distinct nodes $i$ and $j$, there exists a set $Z \subset V \setminus \{i, j\}$ that meets the following conditions: if no element $z \in Z$ is a descendant of any node $w$ on a directed path from $i$ to $j$, or if $z$ is connected to $i$ only by an undirected edge, as long as $Z$ blocks all non-directed paths from $i$ to $j$, then $Z$ is a valid adjustment set for calculating the causal effect $CE(i, j)_G$.

\label{valid Adjustment Sets}
\end{Lemma}

Clearly, in the causal graph $A$, the parents $P_i$ of node $i$ constitute a valid adjustment set for computing $CE(i, j)$. On the one hand, as there are no cycles in a DAG, $P_i$ cannot be a descendant of any node on a directed path from $i$ to $j$, while in a CPDAG, $z\in P_i$ only points to $i$, or is connected to $i$ through an undirected edge. On the other hand, all non-directed paths (confounding paths) from $i$ to $j$ take the form $i\leftarrow \cdots \leftarrow k \to \cdots \to j$. Therefore, all non-directed paths necessarily pass through the parents of $i$, implying that controlling for $P_i$ can block all non-directed paths.

\begin{Definition}[Equivalent Definition of CED]
The Equivalent Definition of CED is
\begin{equation}
    CED(G, A) = \#\begin{Bmatrix}(i, j), i\neq j|\begin{aligned}
	True  \quad\quad\quad\quad & \text{ if } \mathcal{G}(i, j) \neq \mathcal{A}(i, j)\\
	 P_i \text{ does not satisfy Lemma~\ref{valid Adjustment Sets} for } G & \text{ if } \mathcal{G}(i, j) = \mathcal{A}(i, j)
\end{aligned}\end{Bmatrix}.
\end{equation}
\end{Definition}

On the one hand, the calculation of causal effects is based on causal structures, and using bidirectional edges to encode undirected edges can better model these structures. On the other hand, the computational complexity of CED is quite high, meaning that handling large-scale data will significantly increase the required time and resources. However, by using matrix computation methods, we can greatly reduce time costs and improve computational efficiency. Therefore, we also use matrix computation to obtain CED.

\begin{Proposition}[Matrix Form of CED]
Let $G$ and $A$ are two DAGs or CPDAGs with the same number of variables, then the equivalent definition of CED is
\begin{equation}
\begin{aligned}
    CED(G, A)& =\Vert\mathcal{G} - \mathcal{A}\Vert_1\\
     + \sum_{(i,j)\in \mathbb{E}}\mathsf{1}&(\mathcal{T}^i[:, i] \circ \mathcal{T}^i[:, j] + \mathcal{H}^i[i, P_i] \circ \mathcal{H}^i[P_i, j] + \sum_{z\in P_i}\mathcal{M}[i,:] \circ \mathcal{M}[:, j] \circ \mathcal{M}[:, z]> 0)
\end{aligned},
\end{equation}
where $\mathbb{E}$ denotes the set of node pairs which satisfy $\mathcal{G}(i, j) = \mathcal{A}(i, j)$, and $\circ$ represents the sum of the element-wise multiplication of vectors. $\mathcal{T}^i$ represents the reachability matrix after controlling $P_i$, $\mathcal{H}^i$ represents the reachability matrix after opening the collision structure blocked by $P_i$, and $\mathcal{M}^i$ represents the reachability matrix with the paths originating from $j$ closed.
\label{Proposition:CED}
\end{Proposition}

All the proofs are provided in the appendix.

\section{Open causal discovery benchmark (OCDB)}
In this section, we describe how \benchmarkname{} is constructed and how to quickly use it. Moreover, we introduce the real-world datasets, baseline models and evaluation metrics included in \benchmarkname{}. For detailed information on the main interfaces in \benchmarkname{}, please refer to Appendix~\ref{sec:benchmark_details}.
All resources are available at https://anonymous.4open.science/r/OCDB-6B6B.
\label{sec:OCDB}

\begin{figure}[t]
    \centering
    \subfigure[]{\includegraphics[width=0.4\textwidth]{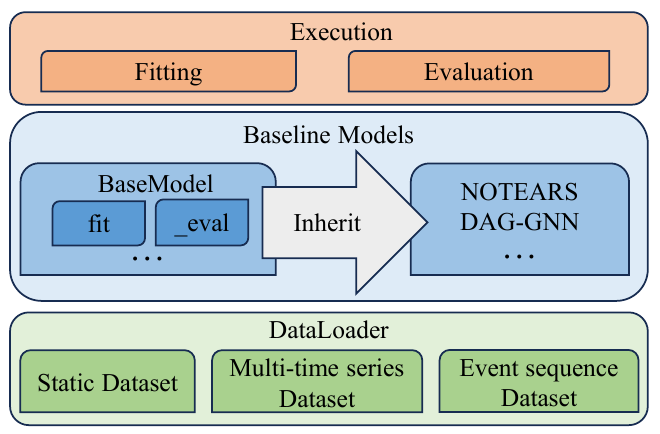}}
    \subfigure[]{\includegraphics[width=0.48\textwidth]{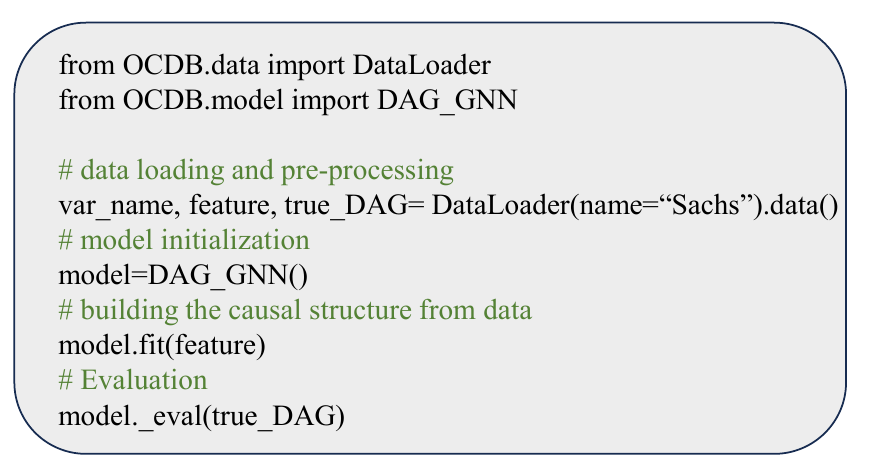}}
    \caption{(a)~The framework of interfaces in \benchmarkname{}. (b)~\benchmarkname{} usage example.}
    \label{fig:framework and usage}
\end{figure}

\subsection{Framework design and implementation}
The framework of \benchmarkname{} interfaces is depicted in Figure~\ref{fig:framework and usage}(a). At the data loading layer, we first curate a diverse and representative set of real-world datasets, covering various domains and data types. These datasets are carefully selected to encompass a wide range of causal relationships and data complexities and are publicly available for research purposes. To ensure ease of use and compatibility, we process and save these datasets in a uniform CSV format and provide the interface \textit{DataLoader} for quick access, manipulation, and analysis of the data.

At the baseline model layer, to facilitate user evaluation and comparison of causal discovery algorithms, we also reconstruct and reproduce several representative baseline models. These models serve as reference implementations that users can use to evaluate their own algorithms or compare against other existing algorithms. By including these baseline models, we provide a standard starting point for users to conduct fair evaluations and performance assessments. To make it easier for users, we offer the class \textit{BaseModel} to standardize and streamline the creation, training, and evaluation of models.

Finally, at the execution layer, the unified interface greatly simplifies the use of \benchmarkname{}. \benchmarkname{} hides all the implementation details, allowing users to easily and quickly perform a series of operations such as accessing and using datasets, creating, training, and evaluating baseline models.

\subsection{Usage of OCDB}
With the support of a unified and comprehensive interface, it becomes straightforward to access the datasets and evaluate the performance of causal discovery algorithms. An example of the simplified usage of \benchmarkname{} is shown in Figure~\ref{fig:framework and usage}(b).
We can easily load datasets using a simple command without caring about how they are processed. Then, we can quickly initialize the desired causal discovery models and use them on the loaded datasets. Finally, using the evaluation methods provided by the interface, we can objectively assess the performance of the algorithm.

This unified interface greatly simplifies the process of dataset loading, algorithm execution, and performance evaluation, allowing researchers to focus on the core aspects of their experiments. It enhances the reproducibility and comparability of results, as all users can follow the same standardized procedures.

\begin{table}[t]
\centering
\caption{The statistics of the datasets included in OCDB.}
\label{tab:Dataset}
\fontsize{9pt}{1pt}\selectfont
\setlength{\tabcolsep}{5pt}
{
\begin{threeparttable}
\begin{tabular}{c|c|c|c|c|c}
\toprule
Category & Dataset & \#Sample & \#Node & \#Edge & Domain \\
\midrule
  & Sachs~\cite{sachs} &  7,466 & 11 & 17 & Bioinformatics\\
 \cmidrule{2-6}
 & DWD~\cite{DWD} & 349 & 5 & 4 & Meteorology\\
 \cmidrule{2-6}
 & Abalone~\cite{Abalone} & 4,177 & 8 & 7 & Bioinformatics\\
 \cmidrule{2-6}
 Static & Auto-mpg~\cite{Auto-mpg} & 398 & 4 & 3 & Mechanical engineering\\
 \cmidrule{2-6}
 & CCS Data~\cite{Css-data} &  1,030 & 9 & 8 & Architecture\\
\cmidrule{2-6}
 & Cad~\cite{Cad} &  450 & 4 & 3 & Pathology\\
 \cmidrule{2-6}
 & Ozone~\cite{Ozone} &  989 & 4 & 3 & Meteorology \\
\midrule[1pt]
 & NetSim~\cite{NetSim} & 10,000 & 15 & 33 & Cognitive neuroscience\\
 \cmidrule{2-6}
 Multi-time series & fMRI-0~\cite{TCDF}  & 21,100 & 220 & 244 & Cognitive neuroscience\\
 \cmidrule{2-6}
 & Finance-8~\cite{TCDF} & 36,000 & 225 & 189 & Economics\\
\midrule[1pt]
 & Wireless~\tnote{*} & 34,838 & 18 & 69 & Industry \\
 \cmidrule{2-6}
 Event sequence & Microwave24V~\tnote{*} & 64,598 & 24 & 137 & Industry \\
 \cmidrule{2-6}
 & Microwave25V~\tnote{*} & 48,572 & 25 & 148 & Industry \\
\bottomrule
\end{tabular}
\begin{tablenotes}
\item[*] https://github.com/gcastle-hub/dataset
\end{tablenotes}
\end{threeparttable}}
\end{table}

\subsection{Real-world datasets, baseline models and evaluation metrics}
For datasets, we collect and curate a total of 13 real-world datasets for causal discovery tasks. Based on the characteristics of the data, they are divided into three categories: 1. Static datasets, which do not contain any time information. 2. Multi-time series datasets, where observations are equally spaced in time intervals. 3. Event sequence datasets, where events occur at irregular time intervals. The brief information on the datasets is shown in Table~\ref{tab:Dataset}. By providing such diverse datasets, we aim to support researchers in comprehensively evaluating and comparing the performance of causal discovery algorithms on different types of data. Such data categorization can help researchers to select suitable datasets according to their research needs and conduct more accurate performance evaluations.

For baseline models, over the years, there has been a development of a range of methods for causal discovery on various types of datasets. 
In the benchmark~\benchmarkname{}, we compile and reproduce some representative causal discovery algorithms addressing different types of datasets. The baseline models are shown in Table~\ref{tab:Baseline}. These methods inherit from the class {\it BaseModel} during implementation, allowing users to quickly and easily create instances, train, and evaluate them. The restructuring and implementation of these methods are based on gcastle~\cite{gcastle}, corresponding papers, and open-source code.

For evaluation metrics, \benchmarkname{} includes SHD-C, CSD, SID, and CED to evaluate causal structure differences and causal effect differences respectively, with SHD-C and SID implementations provided by CDT~\cite{CDT}. Additionally, to meet users' specific needs, we also offer classification-based metrics such as F1-score, TPR, and FPR. Users can easily use various metrics to evaluate models through the interface {\it \_eval}.

\begin{wraptable}{O}{0.5\textwidth}
\centering
\caption{Baseline statistics.}
\vspace{10pt}
\label{tab:Baseline}
\fontsize{9pt}{2pt}\selectfont
\setlength{\tabcolsep}{3pt}
{
\begin{tabular}{c|c|c}
\toprule
Data Type & Baseline & Type\\
\midrule
  & ICA-LiNGAM~\cite{ICA-LiNGAM-TPR+FPR+FNR+TNR} &  FCM-based\\
 \cmidrule{2-3}
 & DirectLiNGAM~\cite{DirectLiNGAM} & FCM-based\\
 \cmidrule{2-3}
 & NOTEARS~\cite{NOTEARS-SHD} & Gradient-based\\
 \cmidrule{2-3}
 Static & NOTEARS$+$~\cite{NOTEARS+} & Gradient-based\\
 \cmidrule{2-3}
 & DAG-GNN~\cite{DAGGNN-SHD+FDR} & Gradient-based\\
 \cmidrule{2-3}
 & GraN-DAG~\cite{GranDAG-SHD+SHDC+SID} & Gradient-based\\
 \cmidrule{2-3}
 & GOLEM~\cite{GOLEM-SHD+SID+TPR} & Gradient-based\\
\midrule[1pt]
& TCDF~\cite{TCDF} & Gradient-based \\
 \cmidrule{2-3}
 Multi-time& GVAR~\cite{GVAR-ICLR-ACC+AUC+AUPR} & Gradient-based \\
 \cmidrule{2-3}
 series& NTiCD~\cite{NtiCD-ACC+Recall+Precision+F1} & Gradient-based \\
\midrule[1pt]
 & ADM4~\cite{ADM4-RelErr+RankCorr} & Gradient-based \\
 \cmidrule{2-3}
 Event& RPPN~\cite{RPPN} & Gradient-based \\
 \cmidrule{2-3}
 sequence& THP~\cite{TTPM} & Score-based \\
 \cmidrule{2-3}
 & SHP~\cite{SHP-F1} & Score-based \\
\bottomrule
\end{tabular}}
\end{wraptable}

\section{Experiments} 
In this section, we showcase the performance of the baseline model on various types of real-world datasets, including our proposed CSD and CED metrics, as well as two metrics for comparing DAGs and CPDAGs: the structure error-based metric SHD-C~\cite{GranDAG-SHD+SHDC+SID} and the causal effect error-based metric SID~\cite{SID}. The comparison of experimental results highlights the superiority of CSD and CED. Additionally, we present the changes in CED computation time as the number of variables increases, with detailed results available in Appendix~\ref{sec:scale_ced}. Finally, detailed experimental settings are provided in Appendix~\ref{sec:experiment_setup}.

\subsection{Experimental results}
All experimental results are shown in Tables~\ref{tab:static}. From the experimental results, we can draw five conclusions as following:

Firstly, when comparing DAG and CPDAG, CSD and CED outperform the other two metrics. SHD-C generates the corresponding CPDAG skeleton for a DAG, causing many structural details to be overlooked. This results in many models having the same SHD-C score, making it difficult to judge superiority, as shown in the experimental results of the DWD dataset. SID, by traversing all MECs to obtain a score range, still makes it hard to compare models within that range as demonstrated in the experimental results of the fMRI-0 dataset. 
CSD and CED consider undirected edges in their definitions, naturally supporting comparisons between DAG and CPDAG. Moreover, the evaluation results of them are single values, making it easier to compare models. Implementing a fair comparison between DAG and CPDAG can help us better identify superior causal discovery methods and support the interpretability of LLMs.

\begin{table}[t]
\centering
\caption{The performance of causal discovery methods on \benchmarkname{} is evaluated. The best results are \textbf{bolded}, while underlining indicates \underline{sub-optimal results}.}
\label{tab:static}
\fontsize{9pt}{1pt}\selectfont
\setlength{\tabcolsep}{1.4pt}
{
\begin{tabular}{c|cccc|cccc|cccc}
\toprule
 & SHD-C & CSD & SID & CED & SHD-C & CSD & SID & CED & SHD-C & CSD & SID & CED \\
\midrule[1pt]
Static & \multicolumn{4}{c|}{Sachs} & \multicolumn{4}{c|}{DWD} & \multicolumn{4}{c}{CCS Data}\\
\midrule
ICA-LiNGAM & {\bf \textcolor{gray}{12}} & 17 & \textcolor{gray}{46} & 56 & \textcolor{gray}{8} & 12 & \textcolor{gray}{16} & 19 & \underline{\textcolor{gray}{19}} & 21 & \underline{\textcolor{gray}{26}} & 45 \\
\midrule
DirectLiNGAM & {\bf \textcolor{gray}{12}} & \underline{16} & \textcolor{gray}{50} & \underline{54} & \textcolor{gray}{8} & 9 & \textcolor{gray}{10} & \underline{16} & \textcolor{gray}{34} & 40 & \textcolor{gray}{37} & 71 \\
\midrule
NOTEARS & \underline{\textcolor{gray}{13}} & 19 & \textcolor{gray}{47} & 61 & \textcolor{gray}{9} & 11 & \textcolor{gray}{13} & 18 & \textcolor{gray}{35} & 39 & \textcolor{gray}{28} & 69 \\
\midrule
NOTEARS$+$ & \textcolor{gray}{48} & 52 & {\bf \textcolor{gray}{32}} & 94 & \textcolor{gray}{8} & 9 & \underline{\textcolor{gray}{5}} & 17 & \textcolor{gray}{31} & 35 & {\bf \textcolor{gray}{25}} & 64 \\
\midrule
DAG-GNN & \textcolor{gray}{18} & 19 & \underline{\textcolor{gray}{43}} & 69 & \underline{\textcolor{gray}{7}} & 9 & \textcolor{gray}{7} & 18 & \underline{\textcolor{gray}{19}} & \underline{20} & \underline{\textcolor{gray}{26}} & 47 \\
\midrule
GraN-DAG & \textcolor{gray}{15} & {\bf 15} & \textcolor{gray}{51} & {\bf 53} & \underline{\textcolor{gray}{7}} & \underline{7} & {\bf \textcolor{gray}{4}} & 17 & {\bf \textcolor{gray}{13}} & {\bf 15} & \textcolor{gray}{28} & {\bf 40} \\
\midrule
GOLEM & \textcolor{gray}{16} & 24 & \textcolor{gray}{55} & 86 & {\bf \textcolor{gray}{6}} & {\bf 6} & {\bf \textcolor{gray}{4}} & {\bf 13} & \textcolor{gray}{20} & 24 & \textcolor{gray}{41} & 59 \\
\midrule[1pt]
Time-Series & \multicolumn{4}{c|}{NetSim} & \multicolumn{4}{c|}{fMRI-0} & \multicolumn{4}{c}{Finance-8}\\
\midrule
ICA-LiNGAM & \textcolor{gray}{36} & 58 & \textcolor{gray}{144} & 166 & \underline{\textcolor{gray}{5}} & \underline{6} & \textcolor{gray}{15} & 18 & \textcolor{gray}{176} & 178 & {\bf \textcolor{gray}{205}} & \underline{493} \\
\midrule
DirectLiNGAM & \underline{\textcolor{gray}{34}} & 58 & \textcolor{gray}{151} & \underline{181} & {\bf \textcolor{gray}{3}} & 8 & \textcolor{gray}{19} & 20 & \textcolor{gray}{170} & 179 & \underline{\textcolor{gray}{301}} & 526 \\
\midrule
TCDF & {\bf \textcolor{gray}{24}} & {\bf 33} & \textcolor{gray}{99} & {\bf 99} & \underline{\textcolor{gray}{5}} & {\bf 5} & \textcolor{gray}{13} & {\bf 13} & {\bf \textcolor{gray}{37}} & {\bf 39} & \textcolor{gray}{316} & {\bf 316}\\
\midrule
GVAR & \textcolor{gray}{45} & \underline{56} & \textcolor{gray}{[55,142]} & 210 & \textcolor{gray}{9} & 16 & \textcolor{gray}{[2,18]} & 20 & \underline{\textcolor{gray}{108}} & \underline{128} & \textcolor{gray}{[396,464]} & 593\\
\midrule
NTiCD & \textcolor{gray}{59} & 68 & \textcolor{gray}{[111,125]} & 210 & \textcolor{gray}{8} & 8 & \textcolor{gray}{[11,16]} & \underline{17} & \textcolor{gray}{163} & 196 & \textcolor{gray}{[350,424]} & 600\\
\midrule[1pt]
Event Sequence & \multicolumn{4}{c|}{Wireless} & \multicolumn{4}{c|}{Microwave24V} & \multicolumn{4}{c}{Microwave25V}\\
\midrule
ADM4 & \underline{\textcolor{gray}{70}} & 87 & \textcolor{gray}{[225, 240]} & 250 & \textcolor{gray}{148} & 177 & \textcolor{gray}{[480, 509]} & 565 & \underline{\textcolor{gray}{157}} & 178 & \textcolor{gray}{[440, 543]} & 599\\
\midrule
RPPN & \textcolor{gray}{81} & \underline{83} & \textcolor{gray}{176} & 196 & \underline{\textcolor{gray}{145}} & \underline{154} & \textcolor{gray}{493} & \underline{495} & \textcolor{gray}{159} & \underline{160} & \textcolor{gray}{553} & \underline{555} \\
\midrule
THP & \textcolor{gray}{78} & 87 & \underline{\textcolor{gray}{172}} & {\bf 172} & \textcolor{gray}{155} & 171 & \textcolor{gray}{481} & 498 & \textcolor{gray}{159} & 179 & \textcolor{gray}{511} & 566 \\
\midrule
SHP & {\bf \textcolor{gray}{68}} & {\bf 68} & {\bf \textcolor{gray}{169}} & \underline{173} & {\bf \textcolor{gray}{138}} & {\bf 142} & \textcolor{gray}{488} & {\bf 494} & {\bf \textcolor{gray}{152}} & {\bf 155} & \textcolor{gray}{545} & {\bf 553}\\
\bottomrule
\end{tabular}}
\end{table}
Secondly, sometimes the SID score of the same model is lower than both SHD-C and CED. According to Lemma~\ref{Lemma:causal effect}, the computation of causal effects is influenced by both the structure and the intervention distributions, whereas SID only considers the intervention distribution. This leads to the anomaly where structural differences are greater than the differences in causal effects. Therefore, the causal discovery methods selected based on this criterion may not always enhance the interpretability of LLMs, and could even be harmful. In contrast, CED is derived from the process of computing causal effects, taking into account both structural and intervention distribution differences, thereby avoiding such discrepancies.

Thirdly, it is not always the case that newer models perform better. In particular, for time-series datasets, methods specifically designed for static data sometimes perform even better than those considering time information. 
This indicates a significant gap between synthetic and real-world data. 
Synthetic data is often generated based on certain assumptions or models, which may not fully reflect the complexity and diversity of the real world. 
However, LLMs are trained on real data. Only by evaluating causal discovery methods on real-world data can we better understand the strengths and weaknesses of the model, identify potential issues in practical applications, and provide stronger support for the interpretability and reliability of LLMs. Additionally, this underscores the importance of establishing real-world benchmarks and conducting evaluations. 

Fourthly, the fact that the best model varies across different datasets indicates a weak robustness of causal discovery algorithms. 
This makes it challenging to provide sufficient interpretability for LLMs. We hope to find a method that is robust on a certain type of data, such as static data, so that when integrated into LLMs, it significantly enhances interpretability and reliability. Current methods are overly focused on optimizing for specific data features or problem contexts, leading to poor performance on other datasets. Future causal discovery research needs to include diverse datasets for evaluation and validation to gradually improve generalization capability and robustness.

Finally, we find that current causal discovery models still have substantial room for improvement, regardless of whether it is in terms of causal structure differences or causal effect differences. Therefore, we look forward to further advances and optimizations, which will enable the development of more advanced models in the future, thus providing better support for LLMs.

\section{Conclusion}
This paper discusses the issues surrounding the selection of causal discovery methods for providing interpretability to LLMs. It highlights that current evaluations of causal discovery algorithms are often one-sided and lack assessments using real datasets, potentially overlooking excellent methods. To address these problems, we propose an innovative evaluation framework.
First, we analyze two key factors affecting interpretability: causal structure and causal effect, and derive metrics to measure differences in them. Additionally, we introduce a comprehensive benchmark called \benchmarkname{}, based on real datasets, to ensure fair and accurate evaluation of causal discovery algorithms in real-world scenarios. By offering a standardized open platform and evaluation metrics, we aim to identify more suitable and superior causal discovery algorithms, thereby enhancing the interpretability and trustworthiness of LLMs and promoting their broader and safer application.

% \section*{Acknowledgments}
% This was was supported in part by......

%Bibliography
\newpage
\bibliographystyle{unsrt}  
\bibliography{references}  

\newpage
\appendix
\section{Limitations}
\label{sec:limitation}
This paper has two main limitations. First, the computational complexity of CED is too high. Although current causal graphs are relatively small, as causal discovery algorithms evolve, they will inevitably be applied to datasets with more variables, which will result in significant time costs for CED computation. 
Second, when extending the definition of an effective adjustment set to CPDAGs, the analysis process for Case 3 (if node i has an edge with an undetermined direction and there are directed paths from i to j through this edge) is not entirely rigorous.

\section{Proof of Lemma~\ref{valid Adjustment Sets}}
\label{Proof:lemma_2}
The key to extending CED to CPDAG lies in how to extend Lemma~\ref{valid Adjustment Sets} to CPDAG, that is, to prove that the parents $PA_i$ of node $i$ is also an effective adjustment set in CPDAG. For the two conditions that an effective adjustment set needs to satisfy, $PA_i$ obviously block all non-directed paths. The question is simplified once again to whether $PA_i$ is not a descendant of any intermediate nodes on the directed path from $i$ to $j$.
According to the four principles of deriving causal edge\cite{PC-TPR+FPR+TDR}, if there is a connection between the Markov equivalence classe (MEC) structure and other nodes in CPDAG, the MEC structure only has outgoing edges and no incoming edges. For example, in the MEC structure $i-k-j$, if there is an edge $p\to i$, then the structure must be determined as $i\to k\to j$, otherwise there would be a contradictory structure of $p\to i\leftarrow k$. If there is an edge $q\to k$, the direction of the structure is also determined as $i\leftarrow k\to j$, otherwise, a recognizable collider structure would appear.

\begin{enumerate}
    \item Case 1: If the directions of the edges connected to nodes $i$ can be determined, and the path from $i$ to $j$ includes a MEC structure, the structure must be $i \cdots \leftarrow MEC \to \cdots j$ or $i \cdots \leftarrow MEC_j$, where $MEC_j$ represents the MEC structure that includes node $j$ . In this case, $PA_i$ obviously cannot be a descendant of $i$, thus meeting the second condition.
    \item Case 2: If node $i$ has an edge with an undetermined direction, and there is no directed path from $i$ to $j$ through this edge, the structure can temporarily be represented as $i-k-p \to \cdots \leftarrow \cdots j$ or $i-k \to \cdots \leftarrow \cdots j$. Even if there exists a node $k \in PA_i$ that is both a parent and a descendant of $i$, $k$ is not a descendant of any intermediate nodes on the directed path from $i$ to $j$, so $PA_i$ still satisfies the second condition of a valid adjustment set.
    \item Case 3: If node $i$ has an edge with an undetermined direction, and there are directed paths from $i$ to $j$ through this edge, the structure must be $i-k-j$ or $i-k \to/-p\to \cdots \to j$. In this scenario, due to the use of bidirectional edges to represent uncertainty in CPDAGs, node $k$ can block both potential directed paths and confounding paths, which might not meet the second criterion of the adjustment set. However, it's important to note that in CPDAGs generated by causal discovery algorithms, the number of node pairs with this special structure is very limited. Additionally, in real causal graphs, the probability of occurrence for both types of MECs is the same, meaning that from an expectation standpoint, whether or not node $k$ is used to block directed paths, the probability of correctly identifying causal effects is only 50\%. Therefore, despite some uncertainty and ambiguity, this approach is considered reasonable as it reflects the inherent uncertainty of CPDAGs.
\end{enumerate}

\section{Proof of Proposition~\ref{Proposition:CED}}
\label{Proof:Proposition}
\begin{equation}
    CED(G, A) = \#\begin{Bmatrix}(i, j), i\neq j|\begin{aligned}
	True  \quad\quad\quad\quad & \text{ if } \mathcal{G}(i, j) \neq \mathcal{A}(i, j)\\
	 PA_i \text{ does not satisfy Lemma~\ref{valid Adjustment Sets} for } G & \text{ if } \mathcal{G}(i, j) = \mathcal{A}(i, j)
\end{aligned}\end{Bmatrix}.
\end{equation}
\begin{enumerate}
    \item For $\mathcal{G}(i, j) \neq \mathcal{A}(i, j)$, which indicates an error in the descendant relationship, the number of errors corresponds to the structural differences between the reachability matrices, denoted as 
    \begin{equation}
        SE = \vert\mathcal{G} - \mathcal{A}\vert_1
        \label{euation: DE}
    \end{equation}

    \item For $\mathcal{G}(i, j) = \mathcal{A}(i, j)$, which indicates identical descendant relationships, it is necessary to verify whether the valid adjustment sets $P_i$ on the predicted causal graph $A$ remain effective on the true causal graph $G$. This involves checking whether the valid adjustment sets block all non-directed paths from node $i$ to $j$ and whether they include any descendants of intermediate nodes on the directed path from $i$ to $j$.
    \begin{itemize}
        \item {\bf Verify whether the valid adjustment sets block all non-directed paths}.  
        The non-directed paths between nodes $i$ and $j$ include self-blocking paths and confounding paths. 
        For self-blocking paths,  when we control the valid adjustment set, it might open up collider structures, thereby forming new directed paths. For example, for collider structures $i\to p\to k\leftarrow q\to j$. When we control $k$, it opens up the path from $p$ to $q$, thus forming a new directed path from $i$ to $j$. Therefore, when we control $P_i$, we need to open the corresponding collider structures and check if there is a new directed path from $i$ to $j$ via $P_i$. Assuming $\mathcal{H}^i$ represents the reachability matrix after processing, if there is a newly formed directed path, then we have $ \mathcal{H}^i[i, P_i]\circ \mathcal{H}^i[P_i, j] > 0$.
        For confounding paths, by controlling the valid adjustment set, information cannot flow through these nodes, so we need to remove these nodes from the causal graph. Additionally, we need to further close the paths emanating from $i$ to avoid the influence of spurious confounding paths such as $k\to i\to j$. Assuming the processed reachability matrix is $\mathcal{T}^i$, if there are unblocked confounding paths, then we have $ \mathcal{T}^i[:, i]\circ \mathcal{T}[:, j]> 0$.

        \item {\bf Verify whether the valid adjustment sets include any descendants of intermediate nodes on the directed path from $i$ to $j$}. When any node $k\in V \setminus \{i, j\}$ is both a descendant of node $i$ and an ancestor of node $j$, it necessarily becomes an intermediate node on the directed path from $i$ to $j$. Therefore, it is sufficient to examine whether there is an intersection between these nodes and the ancestors of the effective adjustment set. When calculating the ancestors of the effective adjustment set, it's necessary to block all directed paths passing through $j$, such as $i\to k\to j\to z$. In this case, $z$ is a descendant of $k$, but $z$ has no impact on calculating the causal effect between $i$ and $j$, so $z$ can still be part of the effective adjustment set. Assuming the processed reachability matrix is $\mathcal{M}^i$, if the effective adjustment set does not meet the conditions, then we have
        $ \sum_{z\in PA_i}\mathcal{M}^i[i,:] \circ \mathcal{M}^i[:, j] \circ \mathcal{M}^i[:,z] > 0$. 
    \end{itemize}
So the difference in intervention distributions can be represented as 
\begin{equation}
    IDE = \sum_{(i,j)\in \mathbb{E}}\mathsf{1}(\mathcal{T}^i[:, i] \circ \mathcal{T}^i[:, j] + \mathcal{H}^i[i, P_i] \circ \mathcal{H}^i[P_i, j] + \sum_{z\in P_i}\mathcal{M}[i,:] \circ \mathcal{M}[:, j] \circ \mathcal{M}[:, z]> 0)
\end{equation}
\end{enumerate}

In summary, the matrix representation of CED is 
\begin{equation}
\begin{aligned}
    CED(G, A)& =\Vert\mathcal{G} - \mathcal{A}\Vert_1\\
     + \sum_{(i,j)\in \mathbb{E}}\mathsf{1}&(\mathcal{T}^i[:, i] \circ \mathcal{T}^i[:, j] + \mathcal{H}^i[i, P_i] \circ \mathcal{H}^i[P_i, j] + \sum_{z\in P_i}\mathcal{M}[i,:] \circ \mathcal{M}[:, j] \circ \mathcal{M}[:, z]> 0).
\end{aligned}
\end{equation}

\section{Pseudocode for calculating CED in Proposition~\ref{Proposition:CED}}
\label{sec:alg}
In this section, we demonstrate how to obtain the CED through matrix operations using pseudocode, as detailed in Algorithm~\ref{alg:1}.
\begin{algorithm}[!ht]
\renewcommand{\algorithmicrequire}{\textbf{Input:}}
\renewcommand{\algorithmicensure}{\textbf{Output:}}
	\caption{Computing causal effect distance}
	\label{alg:1}
	\begin{algorithmic}[1]
		\REQUIRE The adjacency matrix $\mathbf{G}$ of the real causal graph $G$, the adjacency matrix $\mathbf{A}$ of the predicted causal graph $A$, and the number of variables $N_v$.
		\ENSURE The causal effect distance between $G$ and $A$.
            \STATE $\mathcal{G} \leftarrow \mathbb{I}(\mathbf{G+I})^{N_v - 1}$  
            \STATE $\mathcal{A} \leftarrow \mathbb{I}(\mathbf{A+I})^{N_v - 1}$ 
            \STATE $CE \leftarrow \vert G - A \vert_1$
            \STATE $\mathbb{E} \leftarrow \{(i, j)|\mathcal{G}(i, j) == \mathcal{A}(i, j) \quad and \quad i \neq j\}$
            \STATE $IDE \leftarrow 0$
            \FOR{$(i, j) \in \mathbb{E}$}
            \STATE $Z \leftarrow P_i \setminus j$
            \STATE \# check whether opens part of the path that was originally blocked after controlling $Z$
            \STATE $\mathbf{H} \leftarrow \mathbf{G}$
            \IF{$|Z| > 0$}
            \FOR{$z \in Z$} 
            \STATE \# open the collider structures related to $Z$
            \STATE $PA \leftarrow$ the parents of node $z$ on the $\mathbf{H}$ 
            \STATE $H[z, PA] \leftarrow 1$
            \ENDFOR
            \ENDIF
            \STATE $H[j, :] \leftarrow 0$
            \STATE $\mathcal{H} \leftarrow \mathbb{I}(\mathbf{H + I})^{N_v - 1}$
            \IF{$\mathcal{H}[i, Z] * \mathcal{H}[Z, j] > 0$}
            \STATE $IDE \leftarrow IDE + 1$
            \STATE Continue
            \ENDIF
            \STATE \# check whether there are still unblocked confounding paths after controlling $Z$
            \STATE $\mathbf{T} \leftarrow \mathbf{G}$
            \STATE $\mathbf{T}[:, Z] \leftarrow 0$, $\mathbf{T}[Z, :] \leftarrow 0$, $\mathbf{T}[i, :] \leftarrow 0$ \# Control $Z$ and close the directed path from $i$
            \STATE $\mathcal{T} \leftarrow \mathbb{I}(\mathbf{T + I})^{N_v - 1}$
            \IF{$\mathcal{T}[:, i] * \mathcal{T}[:, j] > 0$}
            \STATE $IDE \leftarrow IDE + 1$
            \STATE Continue
            \ENDIF

            \STATE \# check whether $Z$ include any descendants of intermediate nodes on the directed path
            \IF{$|Z| > 0$}
            \STATE $\mathbf{M} \leftarrow \mathbf{G}$
            \STATE $\mathbf{M}[j, :] \leftarrow 0$
            \STATE $\mathcal{M} \leftarrow \mathbb{I}(\mathbf{M + I})^{N_v - 1}$
            \STATE $\mathcal{M}[i, i] \leftarrow 0$, $\mathcal{M}[j, j] \leftarrow 0$
            \STATE $error \leftarrow 0$
            \IF{$z \in Z$}
            \STATE $error \leftarrow \mathcal{M}[i, :] * \mathcal{M}[:, j] *\mathcal{M}[:, z] + error$
            \ENDIF
            \IF{$error > 0$}
            \STATE $IDE \leftarrow IDE + 1$
            \STATE Continue
            \ENDIF
            \ENDIF
            \ENDFOR
            \STATE $ced \leftarrow CE + IDE$
		\RETURN $ced$
	\end{algorithmic}  
\end{algorithm}

\section{Scalability of CED}
\label{sec:scale_ced}
\begin{figure}
    \centering
    \includegraphics[width=0.8\textwidth]{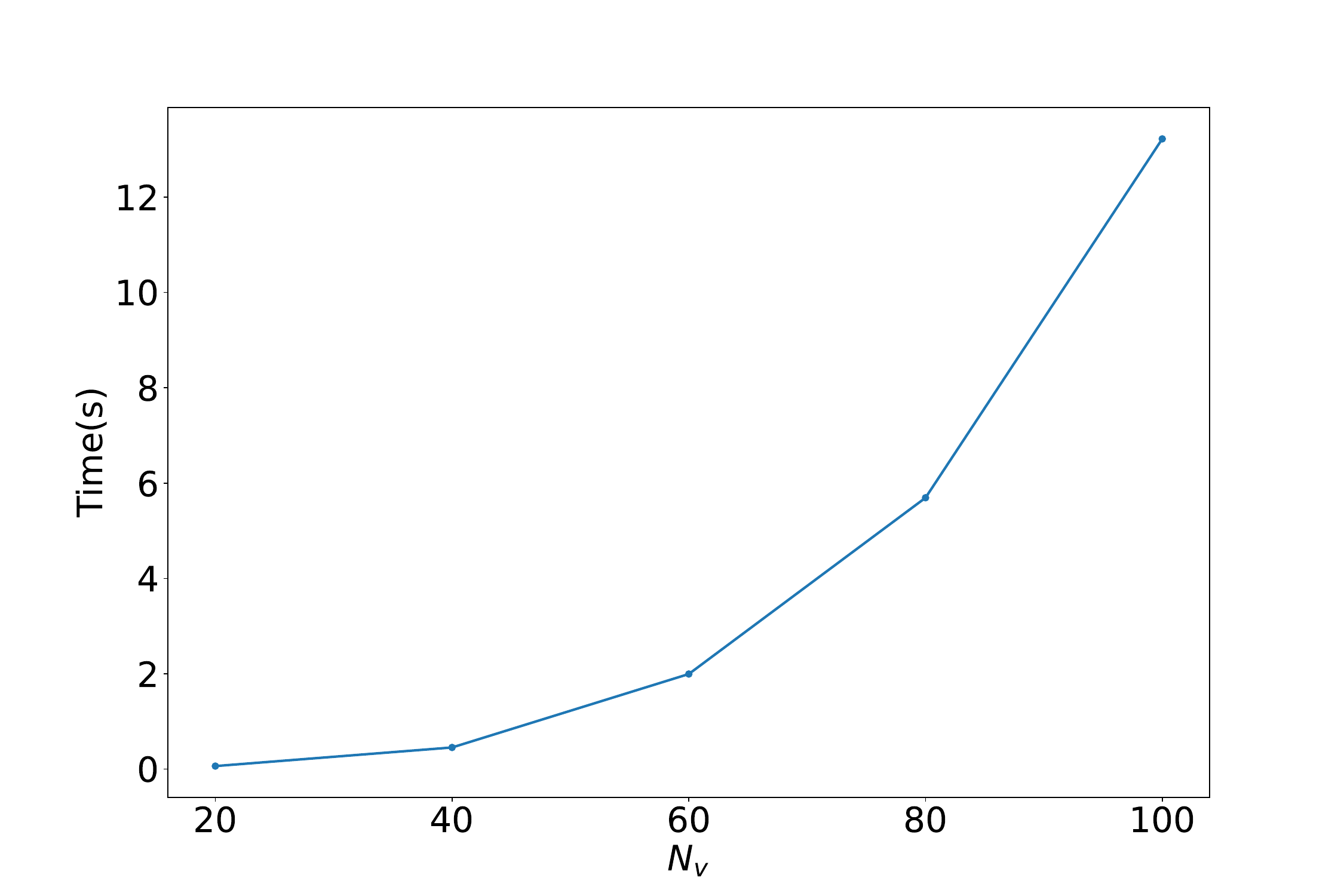}
    \caption{The impact of the number of variables $N_V$ on CED computation time.}
    \label{fig:time_cost}
\end{figure}

In this section, we discuss the impact of the number of variables $N_v$ on the computation time for CED. Specifically, we randomly generate two causal graphs based on the number of variables and set the number of causal edges in these graphs to account for 10\% of all edges. Then, we demonstrate the time cost of computing the CED for these two causal graphs, as shown in Figure~\ref{fig:time_cost}. From the figure, we find that the computational complexity of CED is roughly quadratic or cubic in terms of the number of variables, and the actual time cost is negligible compared to causal discovery algorithms.

\section{Discussion for the current metrics}
\label{sec:discuss}
\begin{table}[ht]
\centering
\caption{Metric statistics.}
\label{tab:Metric}
\fontsize{9pt}{4pt}\selectfont
\setlength{\tabcolsep}{2pt}
{
\begin{threeparttable}
\begin{tabular}{c|c|c}
\toprule
Type & Metric & Calculation \\
\midrule
 \multirow{22}{*}{Structure error-based
} & SHD~\cite{NOTEARS-SHD, NOFEARS-SHD, BCDNets-SHD+SHDC} & $FA + FD + FR$  \\
 \cmidrule{2-3}
 & dSHD~\cite{MCD-dSHD} & $FA + FD + 2 * FR$ \\
 \cmidrule{2-3}
 & SHD-C~\cite{BCDNets-SHD+SHDC, GranDAG-SHD+SHDC+SID} & $FA + FD + FR$ \\
 \cmidrule{2-3}
 & HD~\cite{HD} & $FA + FD + 2 * FR$ \\
 \cmidrule{2-3}
 & Edit-distance~\cite{PNL-EditDistance+ReversedEdges} & $FA + FD + FR$ \\
 \cmidrule{2-3}
 & Reversed-edges~\cite{PNL-EditDistance+ReversedEdges} & $FR$ \\
 \cmidrule{2-3}
 & MRE~\cite{graphite-NLL+MRE} & $\frac{1}{N^2}*FA + \frac{1}{N^2}*FD + \frac{2}{N^2}*FR$ \\
 \cmidrule{2-3}
 & RelErr~\cite{ADM4-RelErr+RankCorr} & $FA + FD + 2 * FR$ \\
  \midrule[1pt]
\multirow{10}{*}{
Causal effect error-based} & KD~\cite{StagedTree-KD+CID} & $\Vert \mathcal{G} - \mathcal{A} \Vert_1$\\
 \cmidrule{2-3}
 & CBC~\cite{OCDaf-CBC+SHD+SID} & $1 - \frac{1}{|E|}\Vert \mathbb{I}(G + G^T) \otimes (\mathcal{G} - \mathcal{A}) \Vert_1$\\
 \cmidrule{2-3}
 & \multirow{2}{*}{SID~\cite{Gaussian-SHD+SID, GOLEM-SHD+SID+TPR, GranDAG-SHD+SHDC+SID}} & $\sum_{i\in V, j \in V, i \neq j}\mathsf{1}(\mathcal{T}^i[:, i] \circ \mathcal{T}^i[:, j] + \mathcal{H}^i[i, P_i] \circ \mathcal{H}^i[P_i, j] $ \\
  &  & $  + \sum_{z\in P_i}\mathcal{M}[i,:] \circ \mathcal{M}[:, j] \circ \mathcal{M}[:, z]> 0)$ \\
\bottomrule
\end{tabular}
\end{threeparttable}}
\end{table}
We collect 10 causal and several classification metrics, analyzing their objectives.
Before analyzing, we need to define some symbols for ease of subsequent discussions. Graph comparison can be divided into 3 cases: 1. {\it False Addition}($FA$), where edges that do not exist in the true graph are wrongly added to the predicted graph. 2. {\it False Deletion}($FD$), where edges that exist in the true graph are mistakenly deleted from the predicted graph. 3. {\it False Reversal}($FR$), where the direction of edges that exist in the true graph is reversed in the predicted graph. 
\subsection{Comparison with structure error-based metrics}
The calculation of structural differences in causal discovery algorithms relies on encoding the causal graphs. Different encoding methods can capture and represent different structural information, leading to variations in the measured structural differences between different graphs.  The analysis results in Table~\ref{tab:Metric} show that all metrics based on structure errors can be expressed in the form as 
\begin{equation}
SE\text{-}like = \alpha * FA + \beta * FD + \gamma * FR,
\label{equation:HD-like}
\end{equation}
which indicates that the different structure error-based metrics are calculated differently but are essentially the same, that is, counting the number of structure errors.

While our proposed SED is represented as 
\begin{equation}
    SED = FA + FD + 2 * FR.
    \label{equation:SED_essence}
\end{equation}

For metrics based on structural errors, despite the existence of several metrics similar to CSD, CSD is the only one that approaches from the perspective of interpretability, and it is the first to consider interpretability errors. It features comprehensive use case analysis and detailed explanations. Additionally, when CSD is defined, it naturally accounts for undirected edges, ensuring fair comparisons between DAGs and CPDAGs. By applying CSD, we can more accurately evaluate the reliability and practicality of models. CSD helps us not only to deeply understand the limitations of models but also guides us in selecting and optimizing models in real-world applications, such as choosing higher-quality causal graphs and supporting the interpretability of LLMs.

\subsection{Comparison with causal effect error-based metrics}
Based on the analysis of CED and existing metrics, we have successfully integrated metrics with unclear evaluation objectives, such as KD and CBC, into the causal effect error category. Furthermore, all metrics based on causal effect error can be expressed in a unified form as 
\begin{equation}
\begin{aligned}
    CE\text{-}like& = \mathbb{C} + \alpha * \Vert\mathcal{G} - \mathcal{A}\Vert_1\\
     + \beta * \sum_{(i,j)}\mathsf{1}&(\mathcal{T}^i[:, i] \circ \mathcal{T}^i[:, j] + \mathcal{H}^i[i, P_i] \circ \mathcal{H}^i[P_i, j] + \sum_{z\in P_i}\mathcal{M}[i,:] \circ \mathcal{M}[:, j] \circ \mathcal{M}[:, z]> 0)
\end{aligned},
\end{equation}
where $\mathbb{C}$ is a constant. 
As Lemma~\ref{Lemma:causal effect} states, the outcomes of causal effects are influenced by both the structure of causal graphs and the intervention distribution.  However, as shown in Table~\ref{tab:Metric}, the current metrics used to assess differences in causal effects usually focus on only one aspect. Specifically, KD and CBC only address the causal effect differences brought about by structural changes, while SID focuses on those caused by changes in intervention distribution. This single-dimensional focus may lead to an incomplete understanding and evaluation of the complex relationships between causal graphs.

Compared with causal effect error-based metrics, our new metric, CED, effectively addresses this issue. CED is based on causal effect calculations, considering structural changes and intervention distribution variations. It captures subtle changes in causal relationships, offering precise evaluation crucial for decision support in research and applications. CED overcomes existing metrics' shortcomings and advances causal discovery methods.

\subsection{Comparison with classification error-based metrics}
Many works treat the measurement of causal graph structures as a classification task, where the presence of a causal edge is labeled as 0 and the absence is labeled as 1. This enables the use of traditional classification metrics to evaluate the performance of causal discovery algorithms. These metrics are also calculated based on the first-order neighborhood relationships, so they have some correlation with the aforementioned $SE\text{-}like$ metrics. We can express them with $FA$, $FD$, and $FR$ by using the following relationships
\begin{equation}
FP = FA + FR \quad \& \quad
FN = FD + FR.
    \label{equation:RelErr}
\end{equation}

For classification error-based, metrics, they can evaluate causal discovery algorithms but have two main drawbacks. First, not all classification metrics suit causal discovery, as causal graphs have far fewer actual edges than non-existent ones, causing a severe class imbalance. Accuracy is inadequate here, and AUPR is better than AUC. Second, classification metrics reflect model performance but don't show graph structure differences, a core feature of causal graphs. Even with the same classification scores, graph structures may vary in measuring causal effects. Thus, focusing only on edge correctness while ignoring graph structure is unrealistic.
Classification metrics provide some information, but they shouldn't be used alone to measure performance.

\section{Main interfaces in \benchmarkname{}}
\label{sec:benchmark_details}
In this section, we provide a detailed introduction to the main interfaces in \benchmarkname{}.
\paragraph{DataLoader.}
The DataLoader is designed to provide unified management and access to data, including functions such as data download and decompression for data pre-processing. To facilitate usage, these functions are hidden, and users only need to use the {\it data} function to easily and quickly access relevant information about the data. The {\it data} function returns the variable names in the data, the data for constructing causal graphs, and the true causal graph structure.
For static and multi-time series datasets, the data for constructing causal graphs is returned as a {\it numpy.array}. For event sequence datasets, the data is returned as a {\it pandas.DataFrame}. The true causal graph structure for all datasets is returned as a {\it numpy.array}.
\paragraph{BaseModel.}
The BaseModel is the parent class of all baseline models, and it provides two key interfaces: {\it fit} and {\it \_eval}. After instantiating a baseline model, you can call the {\it fit} function to generate the corresponding causal structure based on the given data. Additionally, you can continuously adjust and optimize the model output by setting parameters such as epochs and learning rates. 
The {\it \_eval} function is used to evaluate the performance of the current baseline model. You can obtain the corresponding results by setting the name of the evaluation metric. With these two interfaces, anyone can conveniently and quickly use and evaluate all baseline models without having to worry about their implementation details. Finally, the BaseModel also provides other functionalities such as selecting the computation device and obtaining the causal graph structure generated by the model.

\section{Current benchmarks}
\label{sec:current_benchmark}
In Figure~\ref{tab:benchmark}, we present detailed information for each benchmark, including data types, whether there is a baseline model, the objectives of the evaluation metrics, and the time they are proposed.

\begin{table}[t]
\centering
\caption{Benchmark statistics. "S" is short for Synthesis and "R" denotes Real.}
\label{tab:benchmark}
\fontsize{9pt}{4pt}\selectfont
\setlength{\tabcolsep}{2.5pt}
{
\begin{tabular}{c|c|c|c|c|c|c|c|c|c|c}
\toprule
\multirow{2}{*}{Benchmark} & \multicolumn{2}{c|}{Static} & \multicolumn{2}{c|}{Multi-time Series} & \multicolumn{2}{c|}{Event Sequence} & \multirow{2}{*}{Baseline} & \multicolumn{2}{c|}{Metric} & \multirow{2}{*}{Year}\\
\cmidrule{2-7} \cmidrule{9-10}
& Data & DAG & Data & DAG & Data & DAG & & Structure &  Intervention& \\
\midrule
bnlearn & S & R & \XSolidBrush & \XSolidBrush & \XSolidBrush & \XSolidBrush & \Checkmark & \XSolidBrush & \XSolidBrush & 2010\\
\midrule
CauseMe & \XSolidBrush & \XSolidBrush & R & R & R & R & \XSolidBrush & \XSolidBrush & \XSolidBrush & 2019\\
\midrule
CDT & S,R & S,R & \XSolidBrush & \XSolidBrush & \XSolidBrush & \XSolidBrush & \Checkmark & \Checkmark & \Checkmark & 2019\\
\midrule
py-causal & S & S & \XSolidBrush & \XSolidBrush & \XSolidBrush & \XSolidBrush & \Checkmark & \XSolidBrush & \XSolidBrush & 2019\\
\midrule
CDML & \XSolidBrush & \XSolidBrush & S & S & \XSolidBrush & \XSolidBrush & \Checkmark & \Checkmark & \XSolidBrush & 2021\\
\midrule
gCastle & S,R & S,R & S & S & S,R & S,R & \Checkmark & \Checkmark & \XSolidBrush & 2021\\
\midrule
CSuite & S & S & \XSolidBrush & \XSolidBrush & \XSolidBrush & \XSolidBrush & \XSolidBrush & \XSolidBrush & \XSolidBrush & 2022\\
\midrule
CIPCaD-Bench & \XSolidBrush & \XSolidBrush & \XSolidBrush & \XSolidBrush & R & R & \Checkmark & \Checkmark & \XSolidBrush & 2022\\
\midrule
causal-learn & R & R & \XSolidBrush & \XSolidBrush & \XSolidBrush & \XSolidBrush & \Checkmark & \Checkmark & \XSolidBrush & 2023\\
\midrule
CausalTime & \XSolidBrush & \XSolidBrush & R & R & R & R & \Checkmark & \Checkmark & \XSolidBrush & 2023\\
\midrule[1pt]
\benchmarkname{}(Ours) & R & R & R & R & R & R & \Checkmark & \Checkmark & \Checkmark & -\\
\bottomrule
\end{tabular}}
\end{table}

\section{Experimental setup}
\label{sec:experiment_setup}
{\bf Experimental Setting}
The hardware environment for running all baseline models consists of the CPU: Intel Xeon CPU E5-2680, GPU: Tesla P100 (16G), and 250G of running memory. The software environment is Ubuntu 18.04 and CUDA 11.3. The parameters for each model on various datasets are shown below, and the parameters not mentioned are set to default values.

{\bf Sachs}
\begin{itemize}
    \item ICA-LiNGAM: random\_state=2, max\_iter=20, thresh=0.1 
    \item DirectLiNGAM: thresh=0.5  
    \item NOTEARS: w\_threshold=0.5
    \item NOTEARS$+$: max\_iter=5, w\_threshold=0.2, rho\_max=1e5
    \item DAG-GNN: encoder\_hidden=128, decoder\_hidden=128, lr=0.001, epochs=100, k\_max\_iter=20, encoder\_dropout=0.5, decoder\_dropout=0.5, encoder\_type="mlp", decoder\_type="mlp"
    \item GraN-DAG: hidden\_num=1, hidden\_dim=10, batch\_size=64
    \item GOLEM: lambda\_1=0.03, lambda\_2=6, graph\_thres=0.3, num\_iter=20000
\end{itemize}

{\bf DWD}
\begin{itemize}
    \item ICA-LiNGAM: random\_state=42, max\_iter=20, thresh=0.1
    \item DirectLiNGAM:thresh=0.5
    \item NOTEARS:w\_threshold=0.7
    \item NOTEARS$+$:max\_iter=5, w\_threshold=0.5, rho\_max=1e5
    \item DAG-GNN:encoder\_hidden=128, decoder\_hidden=128, lr=0.001, epochs=100, k\_max\_iter=20, encoder\_dropout=0.0, decoder\_dropout=0.0,encoder\_type="mlp", decoder\_type="mlp"
    \item GraN-DAG: hidden\_dim=50, batch\_size=64
    \item GOLEM:lambda\_1=0.02, lambda\_2=6, graph\_thres=0.3, num\_iter=20000
\end{itemize}

{\bf CCS Data}
\begin{itemize}
    \item ICA-LiNGAM:random\_state=2, max\_iter=1000, thresh=1
    \item DirectLiNGAM:thresh=0.0001, measure="kernel"      
    \item NOTEARS:w\_threshold=0.05
    \item NOTEARS$+$:max\_iter=100, w\_threshold=0.3, rho\_max=100
    \item DAG-GNN: encoder\_hidden=256, decoder\_hidden=256, lr=0.001, epochs=50, k\_max\_iter=25, encoder\_dropout=0.5, decoder\_dropout=0.0, graph\_threshold=0.2, encoder\_type="sem", decoder\_type="mlp"
    \item GraN-DAG:hidden\_num=1, hidden\_dim=10, batch\_size=64
    \item GOLEM: lambda\_1=0.02, lambda\_2=5, graph\_thres=0.3
\end{itemize}

{\bf NetSim}
\begin{itemize}
    \item ICA-LiNGAM:random\_state=42, max\_iter=1000, thresh=0.03
    \item DirectLiNGAM: thresh=0.03
    \item TCDF: kernel\_size=128, hidden\_layers=2, threshold=1, epochs=50
    \item GVAR: num\_hidden\_layers=1, hidden\_layer\_size=64, order=6
    \item NTiCD:epochs=20, batch\_size=64, lr=0.001, output\_size=1, hidden\_dim=64, n\_layers=5, window\_size=6
\end{itemize}

{\bf fMRI-0}
\begin{itemize}
    \item ICA-LiNGAM: random\_state=42, max\_iter=100, thresh=0.1
    \item DirectLiNGAM: thresh=0.2
    \item TCDF:  kernel\_size=128, hidden\_layers=2, threshold=1, epochs=10
    \item GVAR: num\_hidden\_layers=1, hidden\_layer\_size=64, order=6
    \item NTiCD:epochs=20, batch\_size=64, lr=0.0001, output\_size=1, hidden\_dim=64, n\_layers=3
\end{itemize}

{\bf Finance-8}
\begin{itemize}
    \item ICA-LiNGAM: random\_state=42, max\_iter=1000, thresh=0.03
    \item DirectLiNGAM: thresh=0.03
    \item TCDF: kernel\_size=128, hidden\_layers=2, threshold=1, epochs=10
    \item GVAR: num\_hidden\_layers=1, hidden\_layer\_size=64, order=6, epochs=20
    \item NTiCD:epochs=10, batch\_size=64, lr=0.001, output\_size=1, hidden\_dim=64, n\_layers=3, window\_size=1
\end{itemize}

{\bf Wireless}
\begin{itemize}
    \item ADM4: graph\_threshold=0.05, max\_iter=10, em\_max\_iter=5, rho=0.2, decay=1
    \item RPPN:embedding\_dim=64, hidden\_size=64, graph\_threshold=0.01, epochs=20, split\_ratio=0.4, init\_scale=2
    \item THP: max\_hop=0
    \item SHP: decay=3, time\_interval=5, seed=42, reg=3, penalty="BIC", threshold=0.5
\end{itemize}

{\bf Microwave24V}
\begin{itemize}
    \item ADM4:graph\_threshold=0.015, max\_iter=10, em\_max\_iter=5, rho=0.2, decay=1
    \item RPPN:embedding\_dim=64, hidden\_size=64, graph\_threshold=0.01, epochs=20, batch\_size=64, split\_ratio=0.5, init\_scale=2
    \item THP: max\_hop = 2
    \item SHP: decay=3, time\_interval=5, seed=42, reg=3, penalty="BIC", threshold=0.9
\end{itemize}

{\bf Microwave25V}
\begin{itemize}
    \item ADM4:graph\_threshold=0.01, max\_iter=10, em\_max\_iter=10, rho=0.2, decay=1
    \item RPPN: embedding\_dim=128, hidden\_size=128, graph\_threshold=0.015, init\_scale=2, epochs=20
    \item THP: max\_hop = 2
    \item SHP: decay=3, time\_interval=5, seed=42, reg=3, penalty="BIC", threshold=0.9
\end{itemize}

\end{document}